\documentclass[10pt,twocolumn,letterpaper]{article}

\usepackage{cvpr}
\usepackage{times}
\usepackage{epsfig}
\usepackage[pagebackref=true,breaklinks=true,letterpaper=true,colorlinks,bookmarks=false]{hyperref}
\usepackage{cleveref}
\usepackage{graphicx}
\usepackage{amsmath}
\usepackage{amssymb}
\usepackage{subcaption}
\usepackage{enumitem}
\usepackage{simplemargins}
\setbottommargin{0.55in}

\usepackage[pagebackref=true,breaklinks=true,letterpaper=true,colorlinks,bookmarks=false]{hyperref}

\cvprfinalcopy % *** Uncomment this line for the final submission

\def\cvprPaperID{15} % *** Enter the CVPR Paper ID here

\ifcvprfinal\fi
\begin{document}

%%%%%%%%% TITLE
\title{Instant Motion Tracking and Its Applications to Augmented Reality}

\author{Jianing Wei,
Genzhi Ye,
Tyler Mullen, \\
Matthias Grundmann,
Adel Ahmadyan,
Tingbo Hou\\
Google Research\\
% 1600 Amphitheatre Pkwy, Mountain View, CA 94043\\
{\tt\small \{jianingwei, yegenzhi, tmullen, grundman, ahmadyan, tingbo\}@google.com}
% For a paper whose authors are all at the same institution,
% omit the following lines up until the closing ``}''.
% Additional authors and addresses can be added with ``\and'',
% just like the second author.
% To save space, use either the email address or home page, not both
}

\maketitle
%\thispagestyle{empty}

%%%%%%%%% ABSTRACT
\begin{abstract}
   Augmented Reality (AR) brings immersive experiences to users. With
   recent advances in computer vision and mobile computing,
   AR has scaled across platforms, and has increased
   adoption in major products. One of the key challenges in enabling AR
   features is proper anchoring of the virtual content to the real world,
   a process referred to as tracking. In this paper, we present a
   system for motion tracking, which is capable of robustly tracking planar
   targets and performing relative-scale 6DoF tracking without calibration.
   Our system runs in real-time on mobile phones and has been deployed in multiple
   major products on hundreds of millions of devices.
\end{abstract}

%%%%%%%%% BODY TEXT
\vspace{-0.5\topsep}
\section{Introduction}
% TODO(jianingwei): fill in this section
% Are mobile phones really faster than desktop computers?
% Also, we probably want to define/explain/reference something for SLAM first.
Mobile phones carry an enormous amount of computational power in a small
package, making them an excellent platform for real-time
computer vision and augmented reality applications. Recent
releases of ARCore \cite{Arcore} and
ARKit \cite{Arkit} scaled Augmented Reality (AR)
to hundreds of millions of mobile devices across major mobile
computing platforms. Their success is built on advances in computer vision,
\eg SLAM \cite{Triggs:1999, Lynen15, Schneider18} and increases in
on-device computational power.

A critical component of AR is the ability to anchor virtual content
to the real world by tracking the environment. Tracking provides the
3D transform that enables the accurate placement and rendering of
virtual content in the real world. Augmentation using virtual
content can be simply overlaying a 2D texture, or
rendering complex 3D characters into real scenes.

In this paper, we propose a novel instant motion tracking system,
based on robust feature
tracking, as well as global and local motion estimation. With a shared
motion analysis module, our system is capable of performing
both planar target tracking and anchor region-based 6DoF tracking
(using a mobile device's orientation sensor).
Unlike SLAM, our system does not require calibration or
initialization to introduce parallax. It is also amenable to tracking
moving regions. By removing the need for calibration,
it enables AR applications to be deployed at a large scale. By removing the need for
initialization, we can place
AR content instantaneously (even on moving surfaces),
without requiring users to translate their phones first.
%Unlike traditional SLAM systems, our system
%does not require calibration, nor initialization. Due to the
%variability of the assorted sensors and cameras in phones,
%SLAM systems require proper initialization and calibrations
%robust performance. Calibration and initialization present
%a huge challenge for shipping AR applications at a large scale,
%across billions of mobile phones. 

Our main contributions are:
\vspace{-0.5\topsep}
\begin{itemize}
\setlength{\parskip}{0pt}
\setlength{\itemsep}{0pt plus 1pt}
    %\item We propose a motion tracking algorithm that estimates
    %relative scale.
    \item A system that is robust in the face of degenerate cases
    like planar scenes, no user motion, and pure rotation.
    \item Calibration-free placement of AR components without a complex
    initialization procedure.
    \item Real-time performance on mobile phones, serving
    millions of users AR content across a wide range of mobile devices.
\end{itemize}

% However, due to the variability of the assorted sensors and cameras in phones,
% SLAM systems require proper initialization and calibrations from the
% manufacturers for
% robust performance. Calibration and initialization present a huge
% challenge for developing AR applications at a large scale,
% across billions of mobile phones.

% TODO: Related work seems to have a lot of information not really directly tied
% to the paper-- is that ok?
\section{Related work}
Standard SLAM pipelines~\cite{Triggs:1999} require users to perform
parallax-inducing motions, the so-called
\emph{SLAM wiggle}~\cite{bustos:2019}, to initialize the
world map~\cite{Mulloni:2013uz}. 
These systems can fail in the presence of degenerate cases,
such as slow-moving cameras,
pure rotations, planar scenes, and tracking distant objects. 
\emph{Homographies}, on the other hand, can accurately and
robustly describe the motion in such cases~\cite{prince:02, Gauglitz:2012}.
%Handling degenerate cases is especially
%important to AR applications because these motions describe
%many common user scenarios, such as sitting still or looking around.

Accurate initialization improves the resilience of SLAM algorithms and makes
optimization converge faster. Researchers have relied on Structure-from-Motion (SfM)
techniques, \eg rotation averaging \cite{bustos:2019,Carlone:wj}, or
closed-form solutions \cite{DominguezConti:2018iy} to initialize the camera
trajectories and the world map. However, these techniques still require
parallax-inducing motion and accurate calibration, rendering them problematic for instant AR placement.

Planar trackers are widely used in SfM applications and
panoramic image registration \cite{Wagner:2010}. \cite{prince:02} studied
planar tracking for augmented reality applications. Direct region tracking
algorithms typically use a homography to warp an image patch from the template
to the source and minimize the difference \cite{Baker:2004}.
\cite{Benhimane:2004} proposed a region-based planar tracker using a
second-order optimization method for minimizing SSD errors.
\cite{Richa:2011} is another region tracker using second-order optimization
to minimize the sum of conditional variances. Lucas-Kanade and compositional
trackers~\cite{Baker:2004} require re-evaluating the Hessian of the loss
function at every iteration. Inverse compositional
trackers~\cite{Baker:2004} speed up tracking by avoiding re-evaluation of the Hessian matrix.

In \cite{Pirchheim:2011}, the authors propose a homography-based planar detection and
tracking algorithm to estimate 6DoF camera poses. Recently, \cite{Arth:2016} used
detected surfaces from an image retrieval pipeline to initialize depth
from the surface map. \cite{Chen:2019} adopted gradient orientation for
direct surface tracking. Correlation filters are also utilized to
estimate rotation as well as scale for 4DoF tracking~\cite{Li:2019}. 
\cite{Liang:2018} built a planar object tracking dataset and surveyed
some of the related work in planar tracking. \cite{Gauglitz:2012} proposed a
model selection algorithm to detect which model, homography, or essential matrix
describes the motion better. %The winning algorithm is then used for tracking.

\section{Instant motion tracking}

\begin{figure}
\hspace{-0.04\textwidth}
\begin{minipage}{0.27\textwidth}
  \centering
\includegraphics[height=1.2\textwidth]{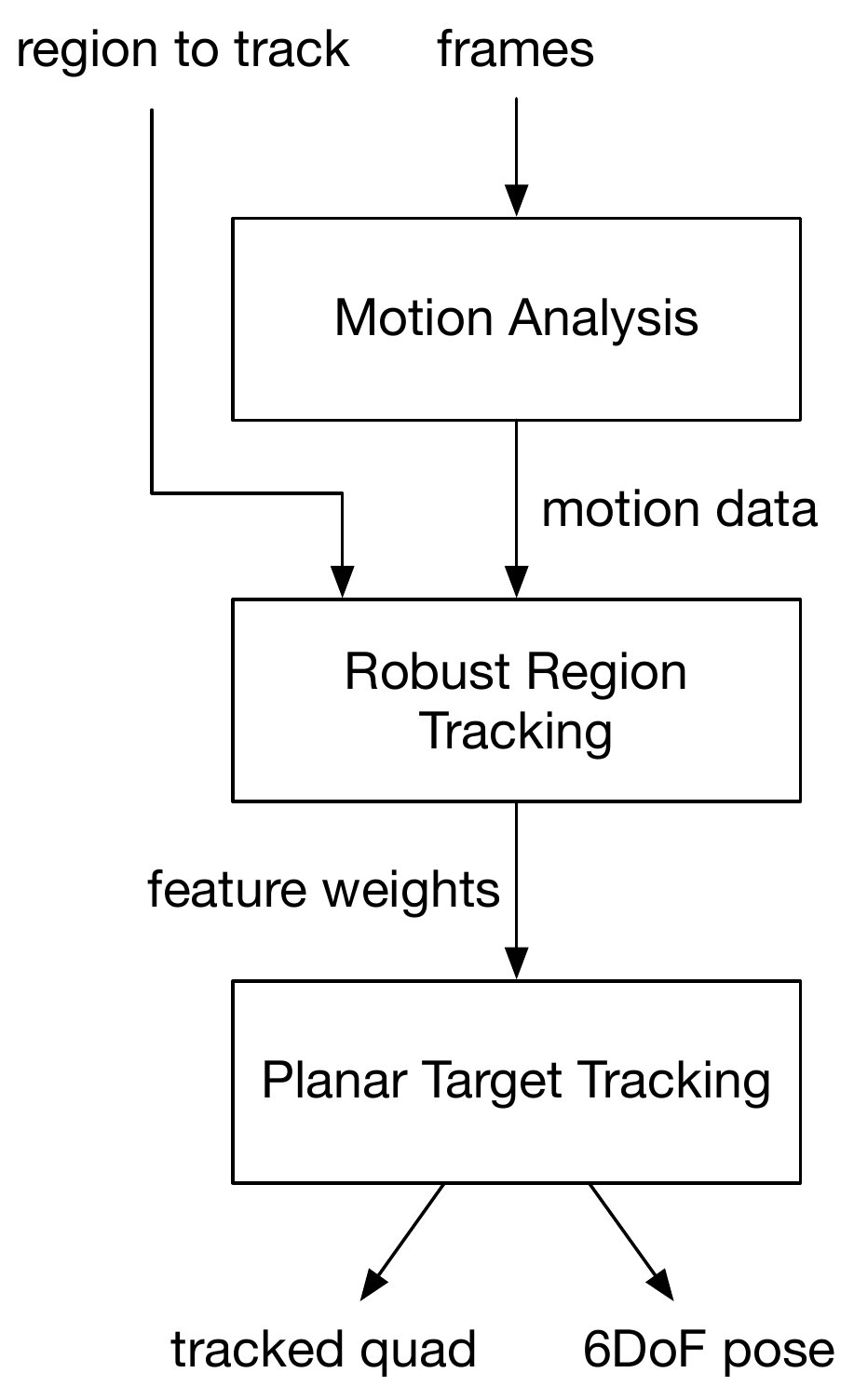}
\subcaption[first caption.]{Planar target tracking}\label{fig:planar_target}
\end{minipage}%
\begin{minipage}{0.27\textwidth}
  \centering
\includegraphics[height=1.2\textwidth]{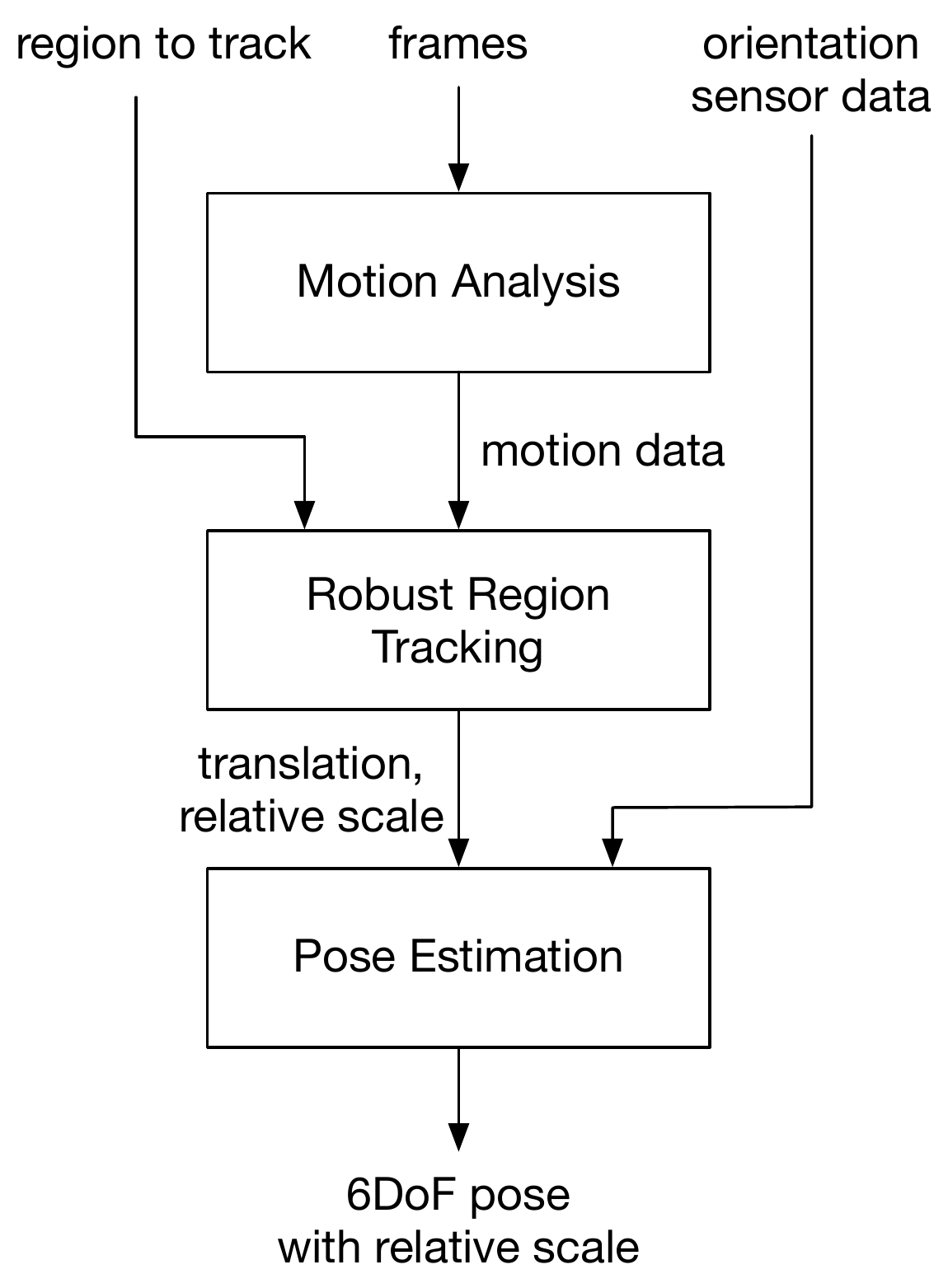}
\subcaption[second caption.]{6DoF tracking}\label{fig:6dof}
\end{minipage}%
\vspace{-6pt}
   \caption{A diagram of our instant motion tracking system.}
\label{fig:pursuit_system}
\end{figure}

Our instant motion tracking system consists of a motion analysis
module, a region tracking module, and either a planar target tracking
module for planar surfaces as shown in fig.~\ref{fig:planar_target},
or a pose estimation module for calibration-free 6DoF tracking as
shown in fig.~\ref{fig:6dof}. In this section,
we will briefly describe each of these modules.

\subsection{Motion analysis}
The motion analysis module extracts and tracks features over time,
classifies them into foreground and background, and estimates a temporally
coherent camera motion model. We create temporally consistent tracks
by assigning each feature path a unique ID. We describe the camera motion by the
highest degree of freedom model that can be robustly computed, depending on number and distribution of features, ranging from a 2DoF
translation model, to a 4DoF similarity model,
and finally to an 8DoF homography model.
The feature extraction and tracking
can be done using standard methods, including finding good features to
track \cite{Shi:1994}.
The feature locations with their IDs and motion vectors (with camera motion subtracted)
for an entire frame, along with the full-frame camera motion model, are packed into
the \emph{motion data} and sent to the region tracking module.

\begin{figure}
\centering
\begin{minipage}{0.24\textwidth}
  \centering
\includegraphics[height=0.58\textwidth]{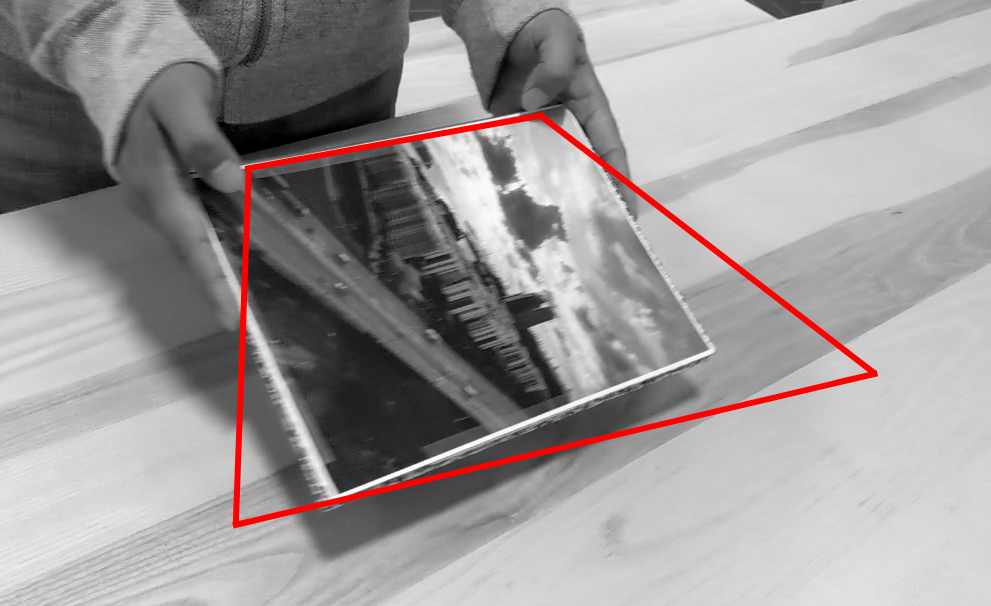}
\subcaption[first caption.]{}\label{fig:homography_perspective_a}
\end{minipage}%
\begin{minipage}{0.24\textwidth}
  \centering
\includegraphics[height=0.58\textwidth]{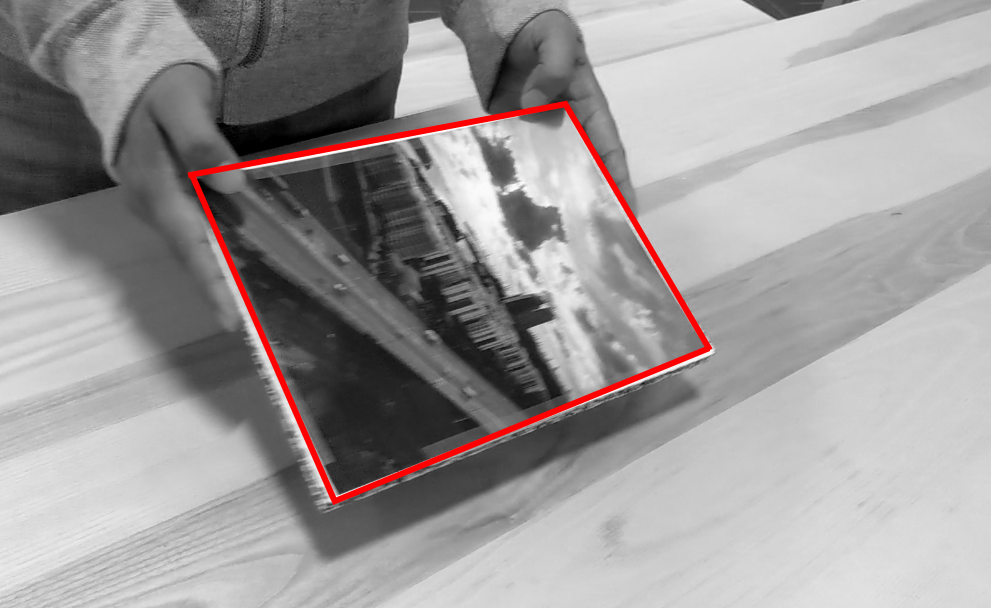}
\subcaption[second caption.]{}\label{fig:homography_perspective_b}
\end{minipage}%
\vspace{-10pt}
   \caption{A comparison between (a) homography tracking and (b) perspective tracking of 500 consecutive frames.}
\label{fig:pursuit_system}
\end{figure}

\subsection{Region tracking}
\label{sec:region_tracking}
% IRLS prior devided by error. where does prior come from
Using solely the motion data produced by the motion analysis module,
the region tracking algorithm tracks individual objects or regions
while discriminating them from others.
%The region tracking module
%can be configured under two settings. For calibration-free
%6DoF tracking, we estimate a 4DoF similarity transform,
%and use the 2D translation and scale in the pose estimation module.
%For planar target tracking, we extend region tracking
%to estimating a 6DoF perspective transform.
%A robust 2D region system is implemented before applying motion
%analysis to full AR effects.
%This places our initial calculations to image space, where we can be
%more confident in accuracy,
%deferring underconstrained 3D computation to the last possible moment.
To track an input region, we first crop the motion data to a corresponding dilated sub-region. Then, using iteratively reweighted least squares (IRLS) we fit a parametric model to the region's weighted motion vectors to determine the region's movement across consecutive frames.

Our importance weights $w_i$ for each vector $v_i$ are of the form $w_i = \tfrac{p_i}{e_i}$ with $p_i$ being the prior of a vector $v_i$'s importance and $e_i$ the iteratively refined fitting error. Each region has a tracking state that defines the prior $p_i$, and includes the mean velocity, the set of inlier and outlier feature IDs, and the region centroid. Note that by relying on feature IDs we implicitly capture the region's appearance since each feature's patch intensity stays roughly constant over time. Additionally, by decomposing a region's motion into that of the camera motion and the individual object motion, we can even track featureless regions.

An advantage of our
architecture is that the motion analysis yields a compact motion metadata over
the full image, enabling great flexibility and constant computation independent of the number of regions tracked. For example, we can easily track multiple
regions simultaneously, and cache metadata across a batch of frames to quickly track regions both backwards and forwards in time; or even sync directly to a specified timestamp for \emph{random access} tracking.

\subsection{Planar target tracking}
\label{sec:planar_target_tracking}
% TODO(yegenzhi): fill in this section
% Planar target tracking and derive 6DoF pose estimation
Many object-centric AR use cases require tracking of planar targets to augment them with virtual objects. 
Examples of planar targets include QR codes, AR markers,
and image/texture targets.
Unlike region tracking (discussed in \cref{sec:region_tracking}),
planar target tracking is capable of providing absolute orientation 
and absolute position--with known physical size.
In this section, we propose a variant of the region tracking algorithm tailored to tracking any planar target 
with known shape (commonly, a rectangle) with high accuracy and resilience.
For the sake of brevity, we limit our discussion to 
a rectangular-shaped planar target.

The goal of planar target tracking is to estimate the image coordinates of the
four corners of the target (a quadrilateral) across frames.
In this scenario, a homography transformation is commonly used to 
describe the inter-frame movement of the quadrilateral \cite{prince:02}. Specifically, a homography matrix is first estimated from
feature correspondences between frames, and applied to update the position of the quadrilateral.
While a homography has 8 degrees of freedom,  in reality, the rigid body transformation that the target undergoes in 3D space is limited to 6 degrees of freedom.
Consequently, the under-constrained nature of the homography transform produces quadrilateral shapes which are not physically possible. These estimation errors, accumulated over time, cause skew artifacts (even disregarding camera lens distortion) as shown in \cref{fig:homography_perspective_a}.

Instead, we advocate using a perspective transform to estimate the
updated corner locations of the quadrilateral. 
Given the 3D coordinates of features in the previous
frame and the corresponding 2D coordinates in the current frame, 
we solve for the rigid body transformation
(3D rotation and translation) from the target local coordinates
to the camera coordinates using Levenberg-Marquardt optimization.
In \cref{fig:homography_perspective_b}, we demonstrate  that our approach reduces skew artifacts and maintains the tracking quality and accuracy
for as long as possible.

\begin{figure}
\begin{center}
   \includegraphics[width=8cm]{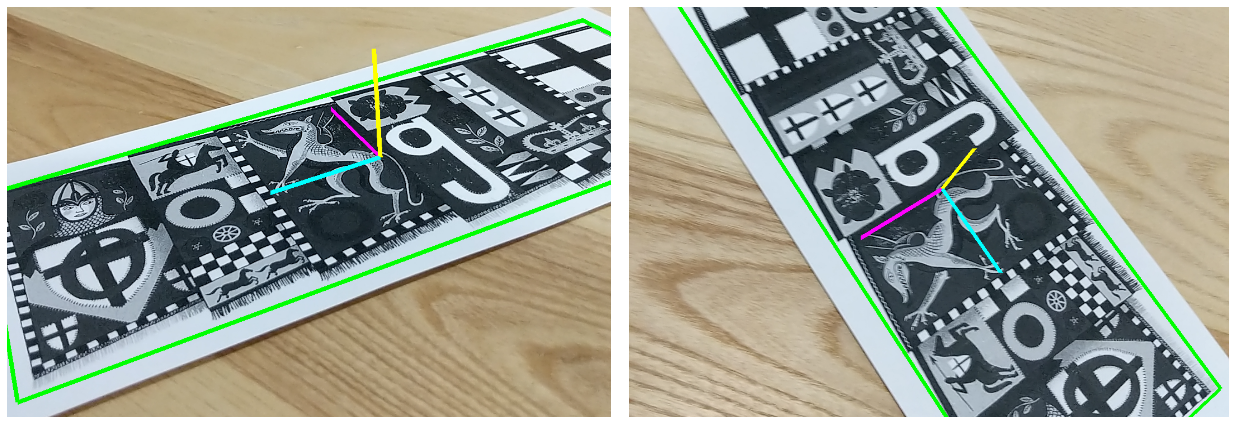}
\end{center}
\vspace{-10pt}
   \caption{Planar target tracking results from two different perspectives, with 3D local coordinates overlaid.}
\label{fig:planar_result}
\end{figure}

\subsection{Pose estimation: calibration-free 6DoF}
% TODO(tmullen, jianingwei): fill in this section
% Add IMU data and you get 6DoF tracking with relative scale
% Emphasize that our system is distinctive from SLAM.
% Advantage: instant placement, calibration free, 
% and tracking of regions whether it's static or moving

% Debating whether this figure is necessary
%\begin{figure}
%\begin{center}
%   \includegraphics[width=4.5cm]{pursuit_scale_estimation.pdf}
%\end{center}
%   \caption{The translation and change in object scale in the
%            image plane can be used to determine 3D translation
%            between two camera positions $C_1$ and $C_2$.}
%\label{fig:pursuit_scale_estimation}
%\end{figure}

The method for planar target tracking described above
is restricted to scenarios with known object geometry.
However, our goal is to enable 6DoF pose estimation
for \textit{all} kinds of targets, to enable users to place 3D virtual objects in the viewfinder, making them appear to be part of the real-world scene. Our key insight to enable this is to decouple the camera's translation and rotation
estimation, treating them instead as independent optimization problems.

We employ our image-based region tracker to estimate translation and relative scale differences. The result models the 3D translation of a tracked
region \wrt the camera (using a simple pinhole camera model).

Separately, the device's 3D rotation (roll, pitch, and yaw) is retrieved from the built-in gyroscope. The local orientation is calculated relative to a canonical global orientation, which we compute on initialization. 
% This local orientation is relativized to a ``global'' space which we create on initialization. 
Using the fused gravity vector from the accelerometer sensor,
the ``up'' direction is observable, % (\ie the orientation of roll and pitch angles),
% The device's accelerometer can tell us which direction is ``up'',
which yields a sensible initial orientation if we further assume initial object placement on a relatively horizontal surface.
%when we further assume that the tracked surface is parallel to the ground plane.
This final assumption is purely based on how we imagine  users will place virtual
objects, and works well in practice.

To make the effect more robust, we also allow for limited region tracking outside
the camera's field of view--enabling a virtual object to
reappear in approximately the same spot when panning away and back again. Combining visual information for 3D translation
and IMU data for 3D rotation lets us track and render virtual content correctly in
the viewfinder with 6DoF, with the initial object scale  being set by the user.
With this parallelization, the system is fast and efficient, can track moving or
static regions, and furthermore requires no calibration--it works on any device with a
gyroscope.

\section{Results and applications to AR}
% TODO(yegenzhi,tmullen,jianingwei): fill in this section

\begin{figure}
\begin{center}
   \includegraphics[width=3.5cm]{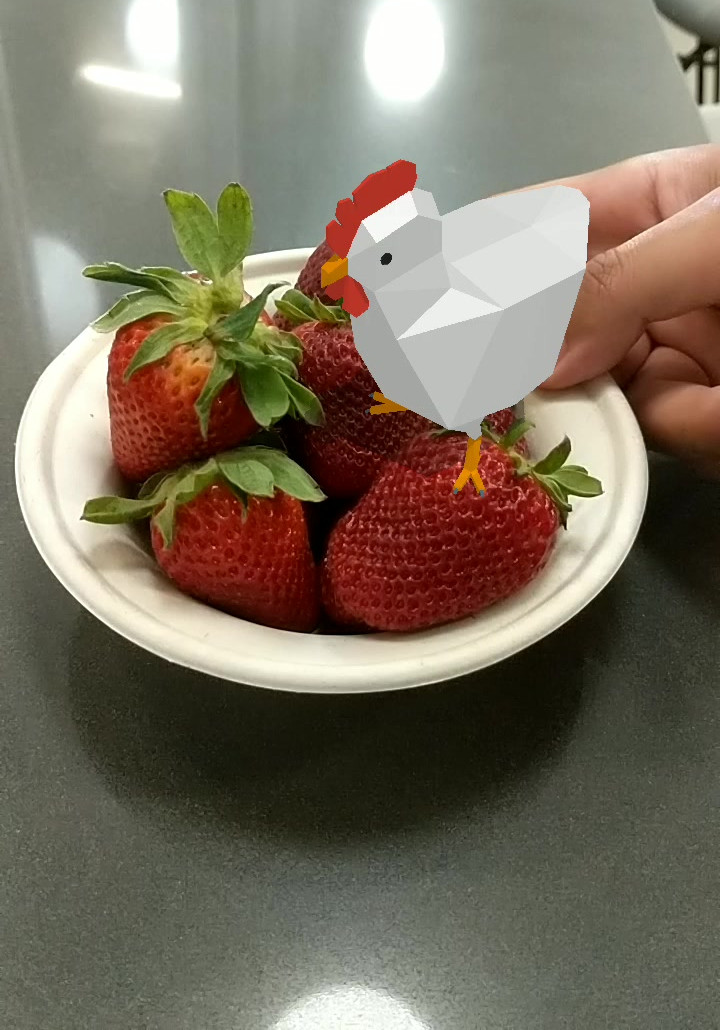}
   \includegraphics[width=3.5cm]{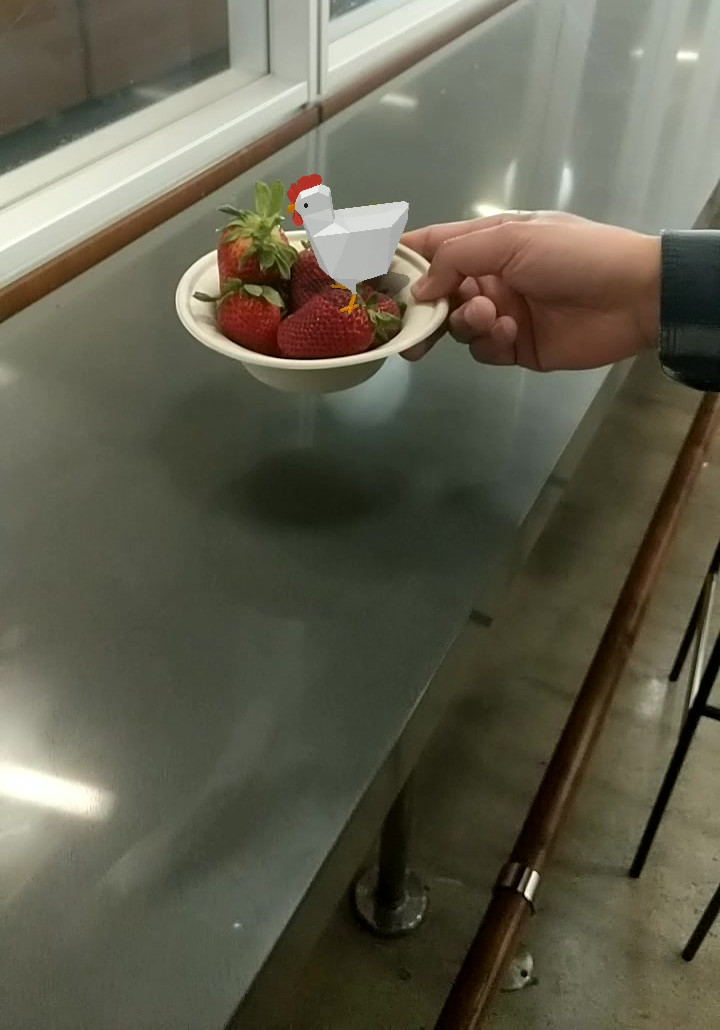}
\end{center}
\vspace{-10pt}
   \caption{AR sticker effect: tracking a moving, non-planar region across a change in camera perspective.}
\label{fig:sticker_result}
\end{figure}

In \cref{fig:planar_result} we demonstrate planar target tracking results for
a real-world image from two different perspectives.
Note that the tracking is capable of producing an accurate 2D quadrilateral
as well as the 3D local coordinate frame of the image. When the physical size of the image is provided, the rotation and translation will be accurate to real-world scale. 
In \cref{fig:sticker_result} we show an AR sticker effect powered by our
calibration-free 6DoF tracking. This technology is driving major AR
self-expression applications on mobile phones, achieving on average 27.5 FPS across 415+ distinct devices.

\section{Conclusion}
% TODO(tingbo,jianingwei): fill in this section
In this paper, we present a system of instant motion tracking, enabling planar target tracking and relative-scale 6DoF tracking. Our system is calibration-free and robust to
degenerate cases. It has been deployed on hundreds of millions of mobile devices, driving major AR applications.
%-------------------------------------------------------------------------

{\small
\bibliographystyle{ieee_fullname}
\bibliography{references.bib}

\begin{thebibliography}{10}\itemsep=-1pt

\bibitem{Arcore}
{ARCore}.
\newblock \url{https://developers.google.com/ar/}.
\newblock [Online; accessed 22-April-2019].

\bibitem{Arkit}
{ARKit}.
\newblock \url{https://developer.apple.com/arkit/}.
\newblock [Online; accessed 22-April-2019].

\bibitem{bustos:2019}
{Alvaro Parra Bustos}, Tat-Jun Chin, Anders Eriksson, and Ian Reid.
\newblock {Visual SLAM: Why Bundle Adjust?}
\newblock {\em arxiv}, pages 1--7, Feb. 2019.

\bibitem{Arth:2016}
Clemens Arth, Christian Pirchheim, Jonathan Ventura, Dieter Schmalstieg, and
  Vincent Lepetit.
\newblock {Instant outdoor localization and SLAM initialization from 2.5 D
  maps}.
\newblock {\em IEEE Transactions on Visualization and Computer Graphics},
  21(11):1309--1318, Nov. 2015.

\bibitem{Baker:2004}
Simon Baker and Iain Matthews.
\newblock Lucas-kanade 20 years on: A unifying framework.
\newblock In {\em International Journal of Computer Vision}, volume~56, page
  221–255, 2004.

\bibitem{Benhimane:2004}
S. {Benhimane} and E. {Malis}.
\newblock Real-time image-based tracking of planes using efficient second-order
  minimization.
\newblock In {\em International Conference on Intelligent Robots and Systems
  (IROS)}, volume~1, pages 943--948 vol.1, Sep. 2004.

\bibitem{Carlone:wj}
L Carlone, R Tron, K Daniilidis, and F Dellaert.
\newblock {Initialization techniques for 3D SLAM: a survey on rotation
  estimation and its use in pose graph optimization}.
\newblock ICRA.

\bibitem{Chen:2019}
Lin Chen, Haibin Ling, Yu Shen, Fan Zhou, Ping Wang, Xiang Tian, and Yaowu
  Chen.
\newblock {Robust visual tracking for planar objects using gradient orientation
  pyramid}.
\newblock {\em J. of Electronic Imaging}, 28(1), Jan. 2019.

\bibitem{DominguezConti:2018iy}
Javier Dom{\'\i}nguez-Conti, Jianfeng Yin, Yacine Alami, and Javier Civera.
\newblock {Visual-Inertial SLAM Initialization: A General Linear Formulation
  and a Gravity-Observing Non-Linear Optimization}.
\newblock {\em IEEE Computer Graphics and Applications}, 2018.

\bibitem{Gauglitz:2012}
S Gauglitz, C Sweeney, and J Ventura.
\newblock {Live tracking and mapping from both general and rotation-only camera
  motion}.
\newblock {\em ISMAR}, pages 13--22, 2012.

\bibitem{Li:2019}
Yang Li, Jianke Zhu, Steven~C.H. Hoi, Wenjie Song, Zhefeng Wang, and Hantang
  Liu.
\newblock Robust estimation of similarity transformation for visual object
  tracking.
\newblock In {\em The Conference on Association for the Advancement of
  Artificial Intelligence (AAAI)}, January 2019.

\bibitem{Liang:2018}
Pengpeng Liang, Yifan Wu, Hu Lu, Liming Wang, Chunyuan Liao, and Haibin Ling.
\newblock Planar object tracking in the wild: A benchmark.
\newblock In {\em Proceedings of the IEEE International Conference on Robotics
  and Automation}, pages 651--658, 2018.

\bibitem{Lynen15}
Simon Lynen, Torsten Sattler, Michael Bosse, Joel Hesch, Marc Pollefeys, and
  Roland Siegwart.
\newblock Get out of my lab: Large-scale, real-time visual-inertial
  localization.
\newblock In {\em Robotics: Science and Systems}, 2015.

\bibitem{Mulloni:2013uz}
Mahesh Ramachandran~Alessandro Mulloni.
\newblock {User Friendly SLAM Initialization}.
\newblock {\em International Symposium on Mixed and Augmented Reality}, pages
  1--10, Oct. 2013.

\bibitem{Pirchheim:2011}
Christian Pirchheim and Gerhard Reitmayr.
\newblock {Homography-based planar mapping and tracking for mobile phones.}
\newblock {\em IEEE International Symposium on Mixed and Augmented Reality},
  pages 27--36, 2011.

\bibitem{prince:02}
Simon~J.D. Prince, {Ke Xu}, and Adrian~David Cheok.
\newblock {Augmented reality camera tracking with homographies}.
\newblock {\em IEEE Computer Graphics and Applications}, 22(6):39--45, Nov.
  2002.

\bibitem{Richa:2011}
R. {Richa}, R. {Sznitman}, R. {Taylor}, and G. {Hager}.
\newblock Visual tracking using the sum of conditional variance.
\newblock In {\em International Conference on Intelligent Robots and Systems},
  pages 2953--2958, Sep. 2011.

\bibitem{Schneider18}
Thomas Schneider, Marcin Dymczyk, Marius Fehr, Kevin Egger, Simon Lynen, Igor
  Gilitschenski, and Roland Siegwart.
\newblock maplab: An open framework for research in visual-inertial mapping and
  localization.
\newblock {\em IEEE Robotics and Automation Letters}, pages 1418--1425, 2018.

\bibitem{Shi:1994}
Jianbo Shi and Carlo Tomasi.
\newblock Good features to track.
\newblock In {\em {Proceedings of IEEE Conference on Computer Vision and
  Pattern Recognition}}, Seattle, WA, USA, 1994.

\bibitem{Triggs:1999}
Bill Triggs, Philip~F. McLauchlan, Richard~I. Hartley, and Andrew~W.
  Fitzgibbon.
\newblock Bundle adjustment - a modern synthesis.
\newblock In {\em Proceedings of the International Workshop on Vision
  Algorithms: Theory and Practice}, ICCV '99, pages 298--372, London, UK, UK,
  2000. Springer-Verlag.

\bibitem{Wagner:2010}
Daniel Wagner, Alessandro Mulloni, Tobias Langlotz, and Dieter Schmalstieg.
\newblock {Real-time panoramic mapping and tracking on mobile phones}.
\newblock {\em 2010 IEEE Virtual Reality Conference (VR)}, pages 211--218, Mar.
  2010.

\end{thebibliography}
}

\end{document}